\documentclass[english]{article}
\usepackage[T1]{fontenc}
\usepackage[latin9]{inputenc}
\usepackage{float}
\usepackage{textcomp}
\usepackage{amsmath}
\usepackage{amsthm}
\usepackage{amssymb}
\usepackage{graphicx}
\usepackage{wasysym}

\makeatletter
\numberwithin{equation}{section}
\numberwithin{figure}{section}
\theoremstyle{plain}
\newtheorem{thm}{\protect\theoremname}
\theoremstyle{remark}
\newtheorem{rem}[thm]{\protect\remarkname}
\theoremstyle{definition}
\newtheorem{defn}[thm]{\protect\definitionname}
\theoremstyle{plain}
\newtheorem{cor}[thm]{\protect\corollaryname}
\theoremstyle{plain}
\newtheorem{lem}[thm]{\protect\lemmaname}

\makeatother

\usepackage{babel}
\providecommand{\corollaryname}{Corollary}
\providecommand{\definitionname}{Definition}
\providecommand{\lemmaname}{Lemma}
\providecommand{\remarkname}{Remark}
\providecommand{\theoremname}{Theorem}

\begin{document}
\title{On the Theory of Bias Tuning in Event Cameras}
\author{David El-Chai Ben-Ezra, Daniel Brisk, Adar Tal}
\maketitle
\begin{abstract}
This paper lays the foundation of a theory for bias tuning in neuromorphic
cameras, a novel sensing technology also known as \textquotedblleft event
cameras''. We begin by formulating the high-level effect of the sensitivity
biases on the camera\textquoteright s event rate in mathematical terms.
We then show that, as a corollary of the Poincaré--Miranda theorem,
the commonly used tuning principles of rate budgeting and polarity
balancing lead to a unique configuration of the sensitivity biases.
As a corollary, we show how by adopting these principles, the multi-variable
bias-tuning problem reduces to a two-parameter problem that can be
resolved experimentally.
\end{abstract}
\\

\section{Introduction}

An event camera is a bio-inspired sensor that represents a paradigm
shift in visual data acquisition. Instead of capturing full frames
at fixed intervals, each pixel operates independently and triggers
only when it detects a brightness change exceeding a threshold. The
output is thus a stream of asynchronous, timestamped events. Event
cameras provide microsecond temporal resolution, low latency, high
dynamic range (over 120 dB), and low power consumption, making them
potentially suited for tasks such as detection, tracking, motion segmentation,
optical flow estimation, 3D reconstruction, pose estimation, and SLAM
\cite{key-1,key-2}, especially in high-speed scenarios.

Because event cameras differ fundamentally from frame-based cameras,
significant effort has been devoted to developing asynchronous, event-driven
processing methods (see \cite{key-3}-\cite{key-24}, \cite{key-25}-\cite{key-28}).
Beyond the output format, they also differ in how data generation
is controlled. Conventional cameras rely on intuitive parameters such
as frame rate and exposure time, guided by well-established principles
like the Nyquist--Shannon theorem. Event cameras, however, do not
use such parameters. Instead, they rely on several biases that control
event generation. Two biases determine pixel sensitivity to illumination
changes, another sets an internal low-pass filter (considered as controlling
the bandwidth \cite{key-1}), a further bias governs an internal high-pass
filter (often overlooked in the literature), and a final bias defines
the pixel refractory period (see $\varoint$\ref{sec:The-biases-of}).
Because these biases directly affect the event stream, proper tuning
is essential for achieving optimal performance. Yet their effects
are less intuitive, and systematic tuning methods remain relatively
immature despite existing studies \cite{key-11}-\cite{key-23}.

A key challenge in developing tuning methods is the lack of direct
information linking bias settings to the generated data. Unlike frame-based
cameras, where photon counts can be inferred from gray levels and
exposure time, an event does not directly indicate the magnitude of
the triggering illumination change. This is due to the logarithmic
sensing mechanism, which depends on the pixel\textquoteright s initial
illumination and varies across pixels. The influence of the internal
filters further complicates this relationship, and the absence of
frames limits intuition. Therefore, there is a need for tuning methods
based on quantitative, user-accessible measurements.

In this paper we adapt two high-level principles proposed for tuning
event camera sensitivity: polarity balancing \cite{key-16} and event-rate
limiting \cite{key-13,key-20}. Since equal sensitivity to increases
and decreases in illumination cannot be set directly, polarity balancing
is typically achieved by matching positive and negative event rates.
Hence, a key advantage of these principles is that they rely on parameters
easily accessible to the user. Event-rate limiting is conceptually
similar to controlling the frame rate in conventional cameras, while
stepping forward and maintaining a constant event rate resembles automatic
gain control (AGC), balancing information capture with preventing
sensor saturation. We emphasize that, throughout this paper, the term
``event rate'' refers to the total number of events generated across
the camera's entire field of view during a scene. 

The paper aims to establish a foundation for event-camera bias tuning,
starting from the aforementioned principles of polarity balancing
and event-rate control, and extending them toward a practical heuristic
for the multi-variable tuning problem. We first describe schematically
how the biases influence the event rate and express these effects
with simple mathematical interpretations (see $\varoint$\ref{sec:Sensitivity-bias-modeling}).
The description is partly based on our experience with the Prophesee
EVK4\nobreakdash-HD, but the underlying properties also follow conceptually
from how event cameras operate. We will also demonstrate the validity
of these properties. However, this paper does not aim to provide a
comprehensive validation of them; rather, it focuses on leveraging
their implications. We say that an event camera that satisfies these
properties in a specific scene has ``standard behavior''. 

Based on this interpretation and the Poincaré--Miranda theorem \cite{key-29},
we derive that given a scene, an event camera with standard behavior,
a choice of the filter biases, and a target event rate $k$, there
exists a unique configuration $(p_{0},n_{0})$ of the sensitivity
biases such that 
\begin{equation}
R_{P}(p_{0},n_{0})=R_{N}(p_{0},n_{0})=k\label{eq:main}
\end{equation}
 where $R_{P}$ and $R_{N}$ are the positive and negative event-rates,
respectively. We note that the theorem only proves the existence of
balanced configuration, and does not relate to the algorithmic way
to converge to it. As a corollary, we obtain that under the aforementioned
assumptions, for any target event rate $k$ and every choice of the
filter biases $l,h$, there exists up to one unique pair $p_{0}(l,h),n_{0}(l,h)$
such that Equation \ref{eq:main} holds.

As a result, under appropriate assumptions about how the target signal
depends on the sensitivity biases, we derive that, under applying
the two high-level principles above, \textbf{the multi-variable problem
of event-camera bias tuning can be reduced to a two-dimensional problem}
that can be explored experimentally. To illustrate this heuristic,
Section $\varoint$\ref{sec:demonstration} analyzes the two-dimensional
case for a periodic light signal driven by the electrical grid. The
analysis shows that the optimal filter bias values differ significantly
from the default settings recommended by Prophesee. In particular,
the proposed heuristic can increase the detected signal by a factor
of 2--4 compared with the default configuration.

Since the goal of this paper is to explain the heuristic rather than
fully validate it, the demonstration does not cover all possible signal
and background conditions. For specific detection tasks, the study
must be adapted accordingly. Broader validation across different scenarios
will be addressed in future work focusing on particular applications.
\begin{rem}
Throughout this paper, the term \textquotedblleft event camera\textquotedblright{}
refers to models that include the tuning parameters discussed above,
excluding devices without these degrees of freedom. For clarity and
concreteness, we focus our discussion on the Prophesee EVK4-HD model.
However, the same heuristic can be adapted to other relevant models
with appropriate adjustments.
\end{rem}

\section{Event Camera Biases\label{sec:The-biases-of}}

Contrary to frame-based cameras, the output of an event camera is
not a sequence of synchronous intensity frames. Instead, it consists
of an asynchronous stream of \textquotedblleft events.\textquotedblright{}

Each pixel maintains a reference intensity level and continuously
monitors changes relative to this reference. When the change in intensity
exceeds a predefined threshold, the pixel updates its reference level
and generates an event. Each event encodes the timestamp (with microsecond
resolution), the pixel coordinates, and the polarity, indicating whether
the change was positive or negative. The camera output is thus a time-ordered
list of such events.

Prophesee event cameras include five parameters that control event
generation, each of which can be tuned upward or downward. Increasing
a parameter in one direction typically leads to more events from the
signal, but also increases background activity and noise; tuning it
in the opposite direction has the reverse effect. We now describe
these parameters.

Two parameters control the thresholds for event generation: one for
positive events and one for negative events. In Prophesee\textquoteright s
interface, these are denoted \textquotedblleft bias\_diff\_on\textquotedblright{}
and \textquotedblleft bias\_diff\_off,\textquotedblright{} which we
refer to as $p$ (positive) and $n$ (negative), respectively. These
parameters can be tuned independently; lowering them reduces the triggering
thresholds, increases sensitivity, and results in more signal events
as well as increased noise.

Two additional parameters control internal low-pass (LPF) and high-pass
(HPF) filters within each pixel. After converting incoming light into
an electrical signal, the pixel passes this signal through both filters
before evaluating whether an event should be generated. These parameters
are labeled \textquotedblleft bias\_fo\textquotedblright{} (LPF) and
\textquotedblleft bias\_hpf\textquotedblright{} (HPF), which we denote
by $l$ and $h$, respectively.

In general, increasing these parameters shifts the frequency response
toward higher frequencies. Increasing $l$ allows more high-frequency
components to pass, leading to more signal events but also higher
background activity. Increasing $h$, on the other hand, suppresses
low-frequency components, reducing both signal events and noise. The
precise relationship between these biases and the effective frequency
response, however, is not well characterized.

The final parameter controls the refractory period of each pixel---the
minimum time between consecutive events. Its main purpose is to suppress
spurious activity caused by internal noise and to limit events from
\textquotedblleft hot pixels,\textquotedblright{} which generate abnormally
high activity. This mechanism is not intrinsic to event-camera operation
and ideally would be unnecessary with perfect pixel stability. In
practice, for the EVK4-HD model, this effect is quite low: the default
refractory period is relatively short (on the order of 10--100 microseconds
\cite{key-17}), and the contribution of hot pixels is negligible.
Therefore, we do not consider this parameter in this work.

We are thus left with four degrees of freedom: two controlling sensitivity
$(p,n)$ and two governing the internal filtering $(l,h)$.

\section{Sensitivity Bias Modeling\label{sec:Sensitivity-bias-modeling}}

We start with some definitions. The term ``scene'' will denote the
environment within the camera\textquoteright s field of view whose
spatial structure, motion, and illumination produce the brightness
variations that trigger events in the sensor. 

The average numbers of positive and negative events generated by scene-induced
brightness variations and internal pixel noise are referred to as
the ``positive event-rate'' and ``negative event-rate'', denoted
by $R_{P}$ and $R_{N}$, respectively. These rates are influenced
by factors such as object motion, flickering light sources, sensor
noise, and small edge displacements caused by air turbulence. Under
this notation, polarity balancing corresponds to the condition $R_{P}=R_{N}$,
while event-rate limiting corresponds to $R_{P},R_{N}\leq k$, for
some predefined bound $k$. The choice of the event-rate bound $k$
should be determined by the user based on various factors, such as
the application requirements, camera stability, computational capacity,
and real-time processing constraints. A detailed characterization
of this bound is beyond the scope of this paper.

Since the biases govern event generation, they also determine the
event rates. Accordingly, we treat $R_{P}(l,h,p,n)$ and $R_{N}(l,h,p,n)$
as functions of the biases. Although, in practice, these parameters
are discrete, we model them as continuous for the purpose of analysis
and assume that the event rates are differentiable with respect to
the biases.

We further assume that increasing sensitivity leads to more information
from the scene, and thus $R_{P}$ and $R_{N}$ are monotonically decreasing
functions of $p$ and $n$. That is, for any choice of $l,h,p,n$ 

\begin{align}
\frac{\partial R_{P}}{\partial p},\frac{\partial R_{N}}{\partial n} & <0\qquad\mathrm{and}\qquad\frac{\partial R_{P}}{\partial n},\frac{\partial R_{N}}{\partial p}\leq0.\label{eq: less}
\end{align}

At first glance, it is not obvious that changing $p$ affects $R_{N}$,
or vice versa. However, lowering a sensitivity threshold has two effects
that increase the event rate. The first is that the camera becomes
responsive to smaller illumination changes. This effect primarily
influences the event rate of the corresponding polarity. In particular,
as a sensitivity bias approaches its minimal value, the corresponding
threshold approaches zero, leading to an unbounded increase in the
associated event rate. Formally, for any $l,h,p,n$

\begin{equation}
\underset{n\to n_{min}}{\lim}R_{N}(l,h,p,n)=\underset{p\to p_{min}}{\lim}R_{P}(l,h,p,n)=\infty\label{eq:infinite.}
\end{equation}
where $p_{min}$ and $n_{min}$ denote the lower bounds of the parameters
$p$ and $n$, respectively.

\begin{figure}[H]
\includegraphics[width=4.8in]{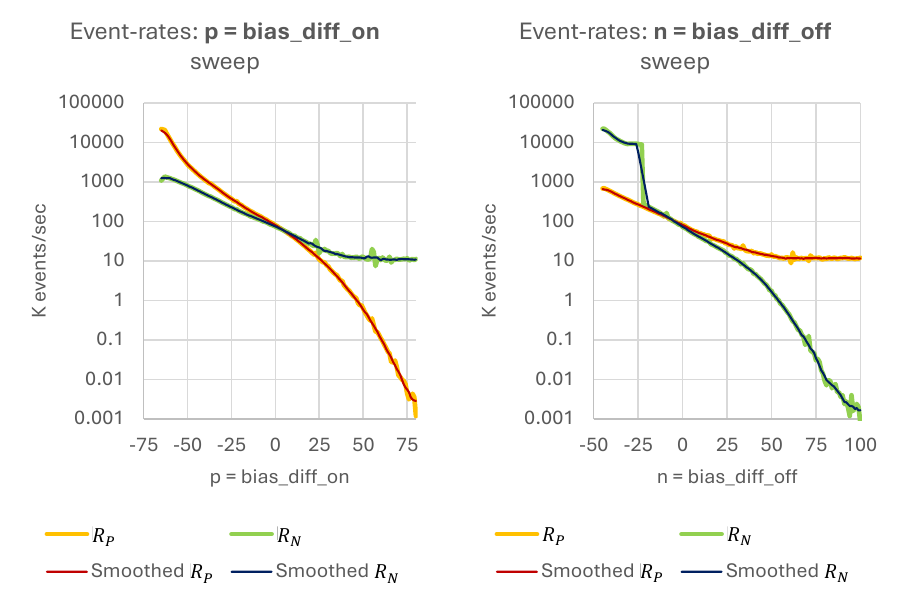}

\caption{Validation of Equations \ref{eq: less} and \ref{eq:infinite.} in
a representative scenario. The event rates $R_{P}$ and $R_{N}$ are
shown as functions of the sensitivity biases, with all other camera
biases fixed at their default values. The left panel presents a sweep
of $p$, while the right panel presents a sweep of $n$. The orange
and green curves depict the positive and negative event rates, respectively.
The red and blue curves show the corresponding seven-sample moving
geometric means, highlighting the underlying trends in the positive
and negative event-rate measurements. \label{fig:rates}}
\end{figure}

Figure \ref{fig:rates} illustrates the above properties in a representative
scenario, where all camera biases were maintained at their default
settings while $p$ and $n$ were varied independently. During the
bias sweeps, we enabled the camera's built-in event-rate cutoff filter,
which limits the total event rate to 20 million events per second
by discarding events that exceed this threshold. The figure presents
the event rates of both polarities as functions of the bias values
on a logarithmic scale, together with smoothed curves obtained by
applying a moving geometric mean with a window size of seven samples.

As shown in the figure, and despite some fluctuations, particularly
at low event rates where statistical variability is more pronounced,
both positive and negative event rates generally decrease monotonically
with increasing values of $p$ and $n$, consistent with the assumption
in Equation \ref{eq: less}. Furthermore, as the sensitivity biases
approach their minimum values, the event rate of the corresponding
polarity increases until it reaches the cutoff limit, in agreement
with the behavior assumed in Equation \ref{eq:infinite.}. When reaching
the cutoff limit, the relative proportion of events with the opposite
polarity decreases, leading to a reduction in their observed event
rate as well. An exception to the general trend can be observed in
the negative event rate during the $n$-sweep, where a noticeable
discontinuity occurs between $n=-23$ and $n=-22$.

The second effect is that lowering a sensitivity threshold allows
the camera to capture a larger fraction of the actual illumination
changes. Unlike the first effect, this influences the event rates
of both polarities. Consequently, reducing even a single sensitivity
bias might increase both $R_{P}$ and $R_{N}$. However, the combination
of both effects is not symmetric. As a rule of thumb, adjusting the
sensitivity of one polarity has a stronger relative impact on the
corresponding event rate than on the opposite one. This can be expressed
as

\begin{align}
\frac{\left(\frac{\partial R_{P}}{\partial p}\right)}{R_{P}} & =\frac{\partial\ln(R_{P})}{\partial p}<\frac{\left(\frac{\partial R_{N}}{\partial p}\right)}{R_{N}}=\frac{\partial\ln(R_{N})}{\partial p}\leq0\nonumber \\
\frac{\left(\frac{\partial R_{N}}{\partial n}\right)}{R_{N}} & =\frac{\partial\ln(R_{N})}{\partial n}<\frac{\left(\frac{\partial R_{P}}{\partial n}\right)}{R_{P}}=\frac{\partial\ln(R_{P})}{\partial n}\leq0.\label{eq:ln-derivatives}
\end{align}

To strengthen the robustness of the analysis, we adopt a weaker assumption.
Specifically, we assume the existence of a monotonic function $\alpha:\mathbb{R}^{+}\to\mathbb{R}$
such that

\begin{align*}
\frac{\partial(\alpha\circ R_{P})}{\partial p} & <\frac{\partial(\alpha\circ R_{N})}{\partial p}\leq0\\
\frac{\partial(\alpha\circ R_{N})}{\partial n} & <\frac{\partial(\alpha\circ R_{P})}{\partial n}\leq0.
\end{align*}

\begin{figure}
\includegraphics[width=4.8in]{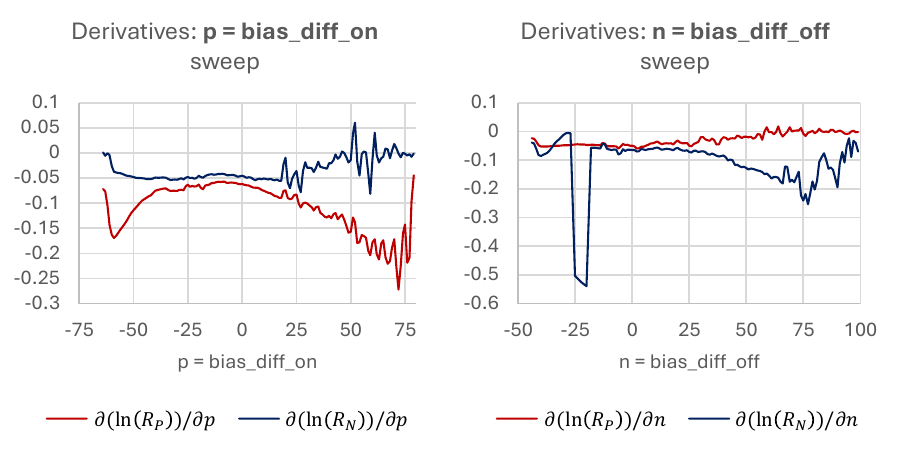}

\caption{Validation of the assumptions in Equation \ref{eq:ln-derivatives}
using the smoothed event-rate curves shown in Figure \ref{fig:rates}.
\label{fig:derivatives}}
\end{figure}

The validity of the assumptions stated in Equation \ref{eq:ln-derivatives}
is demonstrated in Figure \ref{fig:derivatives}, which shows the
derivatives of the logarithm of the smoothed event rates presented
in Figure \ref{fig:rates}. As expected from the decreasing trends
observed in Figure \ref{fig:rates}, the derivatives remain predominantly
negative, except for the previously noted fluctuations in the low-rate
regime, where limited event statistics introduce greater variability.

Moreover, the figure clearly shows that, aside from the previously
discussed discontinuity in the negative event rate during the $n$-sweep,
the corresponding polarity exhibits a larger relative rate of change
than the opposite polarity. This observation is consistent with the
assumption in Equation \ref{eq:ln-derivatives}, indicating that variations
in a given sensitivity bias have a stronger effect on the event rate
of the matching polarity than on that of the opposite polarity.

We summarize the above properties in the following definition.
\begin{defn}
Given a scene, an event camera is said to exhibit ``standard behavior''
if the relationship between the sensitivity biases $p,n$ and the
positive and negative event rates $R_{P}$ and $R_{N}$ satisfies
the following properties for any choice of $l,h,p,n$:
\end{defn}

\begin{enumerate}
\item $\underset{n\to n_{min}}{\lim}R_{N}=\underset{p\to p_{min}}{\lim}R_{P}=\infty$.
\item There exists a monotonic function $\alpha:\mathbb{R}^{+}\to\mathbb{R}$
such that 
\begin{align*}
\frac{\partial(\alpha\circ R_{P})}{\partial p} & <\frac{\partial(\alpha\circ R_{N})}{\partial p}\leq0\\
\frac{\partial(\alpha\circ R_{N})}{\partial n} & <\frac{\partial(\alpha\circ R_{P})}{\partial n}\leq0.
\end{align*}
\end{enumerate}
Our final assumption concerns the dependence of the target signal
on the sensitivity biases. In this context, the ``signal'' refers
to the component of the measured data generated by the phenomenon
of interest, which carries the information to be detected or analyzed.
While the precise definition and measurement of the signal may vary
across applications, it can, in all cases, be modeled as a function
of the biases, denoted by $Sig(l,h,p,n)$. 

We assume that the information captured from the signal increases
with the sensitivity of the camera. Accordingly, we say that a signal
is ``standard'' if, for any $l,h,p,n$

\begin{equation}
\frac{\partial Sig}{\partial p},\frac{\partial Sig}{\partial n}<0.\label{eq:sig}
\end{equation}

\begin{rem}
In general, increasing the signal might also increase the noise around
it, i.e. will also create events around the signal that are not related
to the phenomenon of interest itself. In such a case, one would like
to optimize a certain ``signal-to-noise-ration'' function. As long
as this function behaves ``standard'', we can use it as $Sig$.
In other words, $Sig$ is mostly the function that we want to optimize,
and as long as it has standard behavior, the presented method should
hold.
\end{rem}

\section{The Existence and Uniqueness Theorem\label{sec:The-Existence-and}}
\begin{thm}
\label{thm:main-1}Given a scene, an event camera with standard behavior,
and a choice of the filter biases, let $R_{P}(p,n)$ and $R_{N}(p,n)$
be the positive and negative event-rates as function of the sensitivity
biases $p$ and $n$, respectively. Then, there exists a constant
$K$ such that for every target event-rate $k\geq K$, there exists
a unique pair $(p_{0},n_{0})$ such that 
\begin{equation}
R_{P}(p_{0},n_{0})=R_{N}(p_{0},n_{0})=k.\label{eq:main-1}
\end{equation}
\end{thm}

\begin{rem}
We note that the theorem only proves the existence of balanced configuration,
and does not relate to the algorithmic way to converge to it. 
\end{rem}

\begin{cor}
\label{cor:Given-a-specific-1}Under the assumptions of Theorem \ref{thm:main-1},
given a target event-rate $k$, for every choice of the filter biases
$l,h$, there exists up to one unique pair $p_{0}(l,h),n_{0}(l,h)$
such that Equation \ref{eq:main-1} holds.
\end{cor}

\begin{rem}
In practice, there exist scene conditions and filter-bias configurations
for which the minimal achievable event rate exceeds the desired target
event rate. In such cases, Equation \ref{eq:main-1} admits no solution.
However, under sufficiently aggressive filter-bias configurations,
the lower bound $K$ from Theorem \ref{thm:main-1} is typically close
to zero, indicating that the desired target event rate can still be
achieved.
\end{rem}

Denote the minimal and maximal values of $p$ and $n$ by $p_{min},p_{max},n_{min},n_{max}$.
Given a choice of the filter biases $l,h$, let us denote the following
functions over the region $(-p_{max},-p_{min})\times(-n_{max},-n_{min})$:
\begin{align*}
f(x,y) & =R_{N}(l,h,-x,-y)\\
g(x,y) & =R_{P}(l,h,-x,-y).
\end{align*}
Then, Theorem \ref{thm:main-1} (and Corollary \ref{cor:Given-a-specific-1})
follow from the following lemma.
\begin{lem}
\label{lem:main}Let $f(x,y)$ and $g(x,y)$ be real-valued differentiable
functions, defined over the rectangles $[a,b)\times[c,d]$ and $[a,b]\times[c,d)$,
correspondingly. Assume also that for every $(x_{0},y_{0})$ in the
regions we have:

\begin{align*}
1. & \underset{x\to b}{\lim}f(x,y_{0})=\underset{y\to d}{\lim}g(x_{0},y)=\infty\\
2. & \frac{\partial(\alpha\circ f)}{\partial x}(x_{0},y_{0})>\frac{\partial(\alpha\circ g)}{\partial x}(x_{0},y_{0})\geq0\\
3. & \frac{\partial(\alpha\circ g)}{\partial y}(x_{0},y_{0})>\frac{\partial(\alpha\circ f)}{\partial y}(x_{0},y_{0})\geq0
\end{align*}
for some monotonic $\alpha:\mathbb{R}^{+}\to\mathbb{R}$. Then, there
exist two constants $0\leq K_{1},K_{2}$ such that for every $K_{1}\leq k_{1}$
and $K_{2}\leq k_{2}$ there exists a unique solution to the equation
system
\[
f(x,y)=k_{1},\qquad g(x,y)=k_{2}.
\]
\end{lem}

\begin{proof}
As we assume that $f$ and $g$ are monotonic, we have:
\begin{align*}
K_{1} & :=f(a,d)=\max\{f(a,y)\,|\,y\in[c,d]\}\\
K_{2} & :=g(b,c)=\max\{g(x,c)\,|\,x\in[a,b]\}.
\end{align*}

Now, let $K_{1}\leq k_{1}$ and $K_{2}\leq k_{2}$. Then, by Properties
1-3 there exists $b'\in[a,b)$ and $d'\in[c,d)$ such that
\begin{align*}
\min\{f(b',y)\,|\,y\in[c,d]\}= & f(b',c)>K_{1}\\
\min\{g(x,d')\,|\,x\in[a,b]\}= & g(a,d')>K_{2}
\end{align*}

Thus, by applying Poincaré--Miranda theorem \cite{key-29} on the
rectangle $[a,b']\times[c,d']$, we obtain that there exists a solution
to the equation system
\[
f(x,y)=k_{1},\qquad g(x,y)=k_{2}.
\]

For the uniqueness, notice that using the chain rule, Properties 2
and 3 yield that the of Jacobian of $H:\mathbb{R}^{2}\to\mathbb{R}^{2}$
defined by $H(x,y)=(f(x,y),g(x,y))$ is always positive:
\begin{align*}
J_{H}(x,y) & =\left|\begin{array}{cc}
\frac{\partial f}{\partial x} & \frac{\partial f}{\partial y}\\
\frac{\partial g}{\partial x} & \frac{\partial g}{\partial y}
\end{array}\right|=\frac{\partial f}{\partial x}\cdot\frac{\partial g}{\partial y}-\frac{\partial f}{\partial y}\cdot\frac{\partial g}{\partial x}\\
 & =\frac{1}{\frac{\partial\alpha}{\partial f}\cdot\frac{\partial\alpha}{\partial g}}\left(\frac{\partial(\alpha\circ f)}{\partial x}\cdot\frac{\partial(\alpha\circ g)}{\partial y}-\frac{\partial(\alpha\circ f)}{\partial y}\cdot\frac{\partial(\alpha\circ g)}{\partial x}\right)>0.
\end{align*}
Hence, $h$ is one-to-one, yielding the uniqueness of the solution,
as required. 
\end{proof}
The next lemma shows that, as one should intuitively expect, when
positive and negative event rates are balanced, increasing the target
event rate corresponds to increasing the sensitivity in both polarities.
However, somewhat unexpectedly, as will be remarked later, this relationship
does not necessarily hold when unbalanced sensitivity is required.
\begin{lem}
\label{lem:2}Let $f$ and $g$ be as in Lemma \ref{lem:main}. Then,
both coordinates of the unique solution of the system
\[
f(x,y)=g(x,y)=k
\]
grow as functions of $k$. In other words, $x(k)$ and $y(k)$ are
monotonic.
\end{lem}

\begin{proof}
Let us denote $f_{\alpha}=\alpha\circ f$ and $g_{\alpha}=\alpha\circ g$,
and write the equations
\[
f_{\alpha}(x(k),y(k))=g_{\alpha}(x(k),y(k))=\alpha(k)
\]
where $x(k)$ and $y(k)$ are the unique values given by Lemma \ref{lem:main}
for each $k$. Then, by the chain rule
\[
\begin{array}{c}
\frac{\partial\alpha}{\partial k}=\frac{\partial f_{\alpha}}{\partial k}=\frac{\partial f_{\alpha}}{\partial x}\cdot\frac{\partial x}{\partial k}+\frac{\partial f_{\alpha}}{\partial y}\cdot\frac{\partial y}{\partial k}\\
\frac{\partial\alpha}{\partial k}=\frac{\partial g_{\alpha}}{\partial k}=\frac{\partial g_{\alpha}}{\partial x}\cdot\frac{\partial x}{\partial k}+\frac{\partial g_{\alpha}}{\partial y}\cdot\frac{\partial y}{\partial k}
\end{array}\Rightarrow\left(\begin{array}{cc}
\frac{\partial f_{\alpha}}{\partial x} & \frac{\partial f_{\alpha}}{\partial y}\\
\frac{\partial g_{\alpha}}{\partial x} & \frac{\partial g_{\alpha}}{\partial y}
\end{array}\right)\cdot\left(\begin{array}{c}
\frac{\partial x}{\partial k}\\
\frac{\partial y}{\partial k}
\end{array}\right)=\left(\begin{array}{c}
\frac{\partial\alpha}{\partial k}\\
\frac{\partial\alpha}{\partial k}
\end{array}\right)
\]
yielding that
\[
\left(\begin{array}{c}
\frac{\partial x}{\partial k}\\
\frac{\partial y}{\partial k}
\end{array}\right)=\frac{1}{\frac{\partial f_{\alpha}}{\partial x}\cdot\frac{\partial g_{\alpha}}{\partial y}-\frac{\partial f_{\alpha}}{\partial y}\cdot\frac{\partial g_{\alpha}}{\partial x}}\cdot\left(\begin{array}{cc}
\frac{\partial g_{\alpha}}{\partial y} & -\frac{\partial f_{\alpha}}{\partial y}\\
-\frac{\partial g_{\alpha}}{\partial x} & \frac{\partial f_{\alpha}}{\partial x}
\end{array}\right)\cdot\left(\begin{array}{c}
\frac{\partial\alpha}{\partial k}\\
\frac{\partial\alpha}{\partial k}
\end{array}\right)
\]
and therefore, by the assumptions $\frac{\partial g_{\alpha}}{\partial y}>\frac{\partial f_{\alpha}}{\partial y}\geq0$
and $\frac{\partial f_{\alpha}}{\partial x}>\frac{\partial g_{\alpha}}{\partial x}\geq0$
we have
\begin{align*}
\frac{\partial x}{\partial k} & =\frac{\frac{\partial\alpha}{\partial k}\cdot\left(\frac{\partial g_{\alpha}}{\partial y}-\frac{\partial f_{\alpha}}{\partial y}\right)}{\frac{\partial f_{\alpha}}{\partial x}\cdot\frac{\partial g_{\alpha}}{\partial y}-\frac{\partial f_{\alpha}}{\partial y}\cdot\frac{\partial g_{\alpha}}{\partial x}}>0\\
\frac{\partial y}{\partial k} & =\frac{\frac{\partial\alpha}{\partial k}\cdot\left(\frac{\partial f_{\alpha}}{\partial x}-\frac{\partial g_{\alpha}}{\partial x}\right)}{\frac{\partial f_{\alpha}}{\partial x}\cdot\frac{\partial g_{\alpha}}{\partial y}-\frac{\partial f_{\alpha}}{\partial y}\cdot\frac{\partial g_{\alpha}}{\partial x}}>0
\end{align*}
as required.
\end{proof}
\begin{rem}
One should notice that by following the above argument, it can be
shown that Lemma \ref{lem:2} is not necessarily right under imbalanced
requirement such as $f(x,y)=k$ and $g(x,y)=\gamma\cdot k$ where
$0<\gamma\neq1$. In other words, if a user intentionally seeks an
imbalanced configuration between positive and negative event rates,
achieving a higher total event rate may require the unintuitive adjustment
of one of the sensitivity biases in the opposite direction.
\end{rem}

Lemma \ref{lem:2} implies that any standard signal will grow as function
of the total event-rate, under the principle of polarity balancing,
and standard behavior of the camera:
\begin{cor}
Given a scene, event camera with standard behavior, a standard signal
$Sig(l,h,p,n)$, and filter biases $l,h$, let $p(k)$ and $n(k)$
be the unique solution of the equation $R_{P}(p,n)=R_{N}(p,n)=k$.
Then, $Sig$ is monotonically increasing as function of $k$. 
\end{cor}

\begin{proof}
By Lemma \ref{lem:2}, the chain rule, and Equation \ref{eq:sig},
we have
\[
\frac{\partial Sig(p(k),n(k))}{\partial c}=\frac{\partial Sig}{\partial p}\cdot\frac{\partial p}{\partial k}+\frac{\partial Sig}{\partial n}\cdot\frac{\partial n}{\partial k}>0.
\]
\end{proof}
\begin{cor}
Given a scene, a standard behavior of the camera, a standard signal
$Sig(l,h,p,n)$, filter biases $l,h$, and threshold event rate $k$,
the signal reaches its highest value under the constraints
\begin{align*}
R_{P} & =R_{N}\leq k
\end{align*}
when $p(k),n(k)$ are the unique solutions of the equation $R_{P}=R_{N}=k$. 
\end{cor}

The meaning of the last corollary is that if we balance between bounding
the event rate of the camera, and having as much information from
our signal, then under the constraint of having balanced positive
and negative event-rates, the optimal signal will be reached under
the largest accepted event-rate. In particular, we obtain:
\begin{cor}
\label{cor:Let--be}Let $Sig(l,h,p,n)$ be a standard target signal,
as function of the filter-biases $l,h$ and sensitivity biases $p,n$.
Then, for optimizing the signal under the principles of polarity balancing
and event-rate limiting, it is enough to examine 
\[
\tilde{Sig}(l,h)=Sig(l,h,p_{0}(l,h),n_{0}(l,h))
\]
over the filter-biases $l,h$ plane, where $p_{0}(l,h),n_{0}(l,h)$
are are those biases that solve the system of equations
\[
R_{P}(l,h,p_{0}(l,h),n_{0}(l,h))=R_{N}(l,h,p_{0}(l,h),n_{0}(l,h))=k
\]
given from Corollary \ref{cor:Given-a-specific-1}.
\end{cor}

\section{Determining Filter Biases: A Demonstration\label{sec:demonstration} }

In this section, we present an experimental illustration of the behavior
of the functions $p_{0}(l,h)$, $n_{0}(l,h)$ and $\tilde{Sig}(l,h)$. 

For the signal, we used an incandescent lamp powered by a 50 Hz sinusoidal
electrical grid, producing a 100 Hz periodic signal of the form $|\sin(x)|$.
The signal was measured using a Prophesee EVK4-HD placed a few meters
from the lamp. To simulate a weak signal, the lamp was covered, leaving
only a small aperture for light emission. This setup produced a signal
consisting of a few dozen events per period. The laboratory environment
was kept dark to minimize background activity.

Next, we sampled the $(l,h)$ plane, corresponding to \textquotedblleft bias\_fo\textquotedblright{}
and \textquotedblleft bias\_hpf,\textquotedblright{} respectively,
in steps of 10 units. For each sample, we applied an automatic feedback
control algorithm to balance the positive and negative event rates
while maintaining a total event rate of approximately 100k events
per second (equivalent to positive and negative event-rates of $k$
= 50k event/sec each). From our experience for such scenes, such target
event-rate is higher then the lower bound $K$ required in Theorem
\ref{thm:main-1} for any choice of the filter-biases, and hence convergence
of the algorithm to the desired target event rate is guaranteed. We
remind the reader that the event rates discussed in this paper refer
to the total number of events generated across the camera's entire
field of view during a scene.

\begin{figure}[!ph]
\includegraphics[width=5in,height=7in]{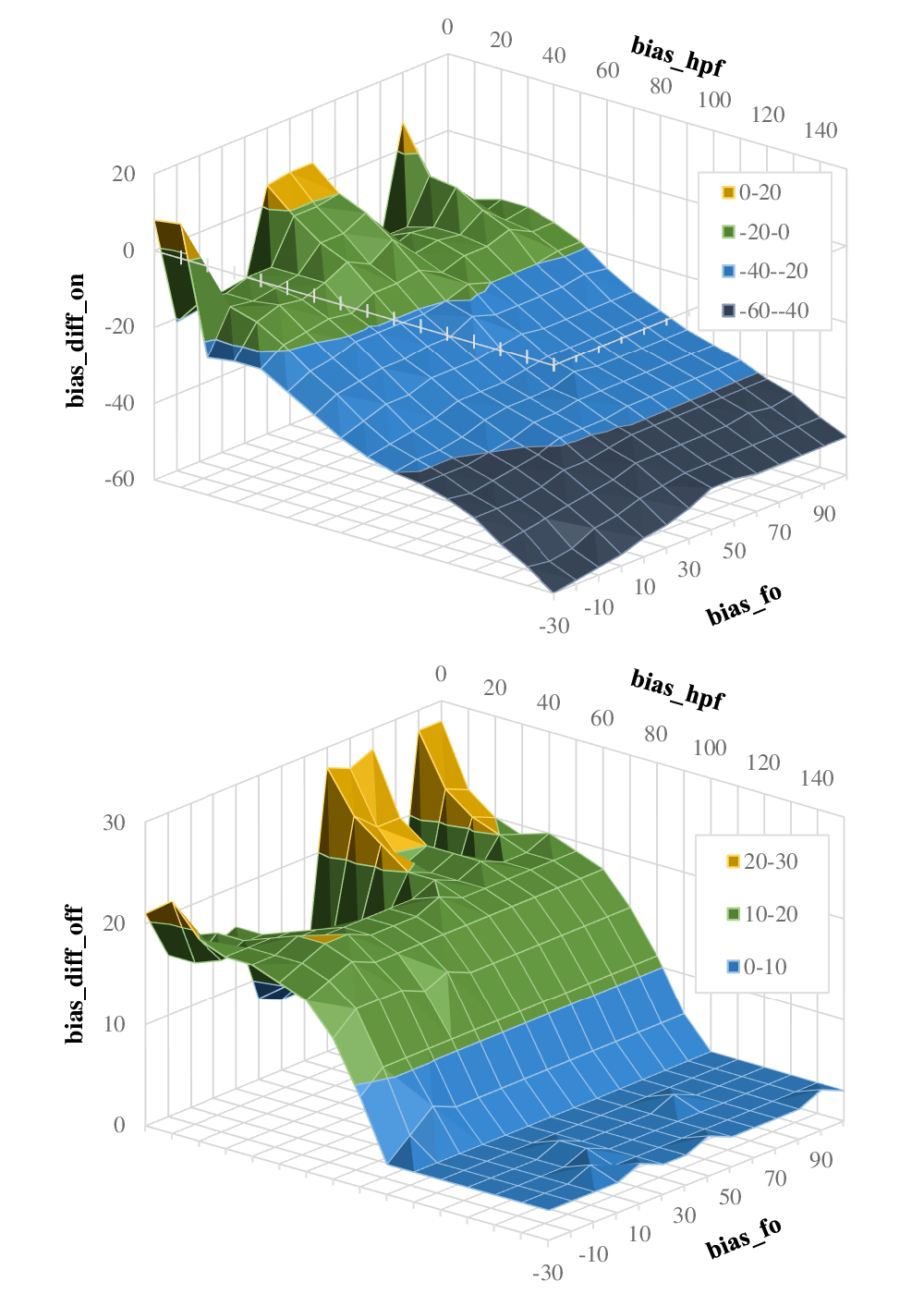}\caption{Values of the sensitivity biases $p$ = bias\_diff\_on and $n$ =
bias\_diff\_off as functions of the filter biases $l$ = bias\_fo
and $h$ = bias\_hpf. For each sampled $(l,h)$, the sensitivity biases
were tuned to maintain a polarity balanced event rate of approximately
100k events/sec.\label{Fig1}}
\end{figure}

In Figure \ref{Fig1}, the values of \textquotedblleft bias\_diff\_on\textquotedblright{}
and \textquotedblleft bias\_diff\_off\textquotedblright , namely $p_{0}(l,h)$
and $n_{0}(l,h)$, are shown for each sample. In general, these values
decrease as \textquotedblleft bias\_hpf\textquotedblright{} increases
and \textquotedblleft bias\_fo\textquotedblright{} decreases. This
indicates that when the filters are more aggressive, the background
noise is reduced, allowing the algorithm to increase the camera\textquoteright s
sensitivity by lowering the sensitivity biases while maintaining a
constant  event rate.

For each sample, after tuning the sensitivity biases, we recorded
the signal for 0.5 seconds, corresponding to 50 periods. We then counted
the number of positive and negative events within a radius of 5 pixels
around the signal center and divided by 50. Accordingly, we define
$\tilde{Sig}(l,h)$ as the average number of events per period in
this region---once for positive events and once for negative events---generated
by the periodic signal of the covered incandescent lamp.

\begin{figure}[H]
\includegraphics[width=5in,height=7in]{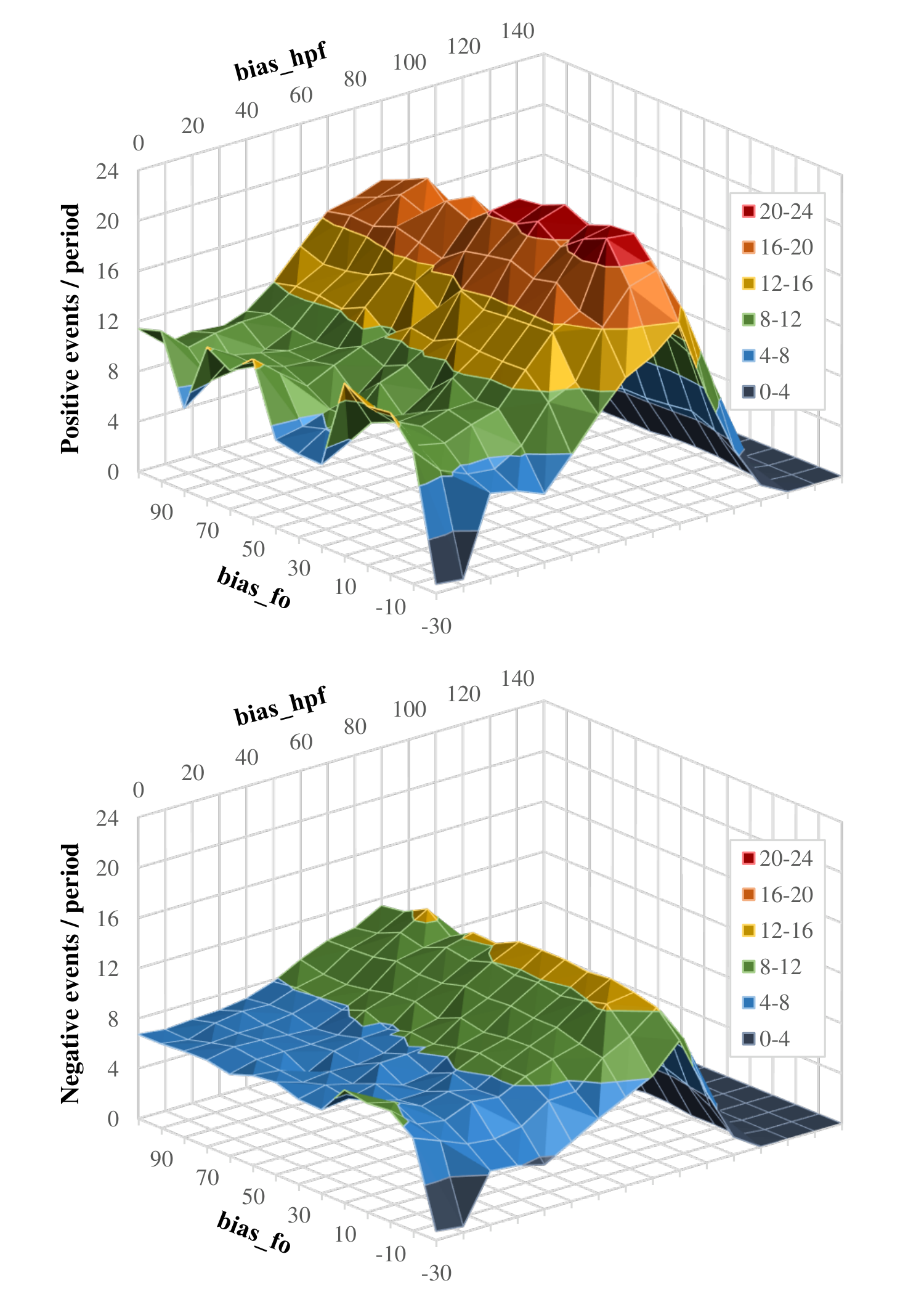}\caption{Number of positive and negative events per period of the signal generated
by the incandescent lamp, as a function of the filter biases.}
\label{Fig2}
\end{figure}

Figure \ref{Fig2} shows the variation of $\tilde{Sig}(l,h)$ as a
function of the filter biases. It can be observed that $\tilde{Sig}$
attains its maximum values for 80$\leq$bias\_hpf$\leq$90 and \textminus 10$\leq$bias\_fo$\leq$0,
in both cases. In particular, when $\tilde{Sig}$ denotes the number
of positive events per period, its optimal value is 2--4 times higher
than the values obtained near the default settings of \textquotedblleft bias\_hpf\textquotedblright{}
and \textquotedblleft bias\_fo\textquotedblright{} (both equal to
0 in the Prophesee EVK4-HD). In the case of negative events, the optimal
value is slightly lower, reaching approximately 1.5--2 times the
default value. This difference between positive and negative responses
highlights the importance of defining $\tilde{Sig}$ appropriately
for the desired task. Additionally, the results indicate that the
signal is significantly more sensitive to \textquotedblleft bias\_hpf\textquotedblright{}
than to \textquotedblleft bias\_fo,\textquotedblright{} a somewhat
surprising observation given the limited attention paid to \textquotedblleft bias\_hpf\textquotedblright{}
in the literature.
\begin{rem}
The tuning algorithm iterates over the biases in strips, fixing \textquotedblleft bias\_fo\textquotedblright{}
while varying \textquotedblleft bias\_hpf\textquotedblright{} in steps
of 10 units, from its minimal value up to bias\_hpf = 150. As a result,
there is a significant difference between the background event rate
at the last sample of one strip and that at the first sample of the
next strip prior to retuning. This discrepancy may cause the algorithm
to converge improperly at the beginning of each strip and explains
the observed valleys at lower values of \textquotedblleft bias\_hpf\textquotedblright .
\end{rem}

\section{Conclusions}

In this paper, we developed a theoretical framework for bias tuning
in event cameras, motivated by the fact that the relationship between
the sensitivity biases and the generated events is often vague, making
direct tuning difficult. To overcome this, we proposed a solution
based on quantities accessible to the user, such as the event rate,
and formalized the principles of polarity balancing and event-rate
control. This approach allows the inherently multi-variable bias tuning
problem to be reduced to a structured, two-parameter problem, making
it both analyzable and experimentally tractable. A key theoretical
insight is that the existence of a balanced bias configuration follows
from the Poincaré--Miranda theorem, demonstrating that bias tuning
is grounded in rigorous mathematical principles rather than heuristics.
The notions of ``standard behavior'' for the camera and ``standard
signal'' provide a formal framework to describe how both background
activity and the target signal depend on the sensitivity biases. Experimental
results with a periodic light source illustrate that the framework
effectively guides tuning, significantly increasing the signal while
controlling background events, and highlight the important role of
the high-pass filter bias, which is often underexplored in the literature.
Overall, this work bridges theory and practice, providing a solid
mathematical basis for bias tuning in event cameras. Future work will
extend this framework to systematic validation across diverse applications.

\section*{Declaration }

During the preparation of this work the authors used AI-assisted technology
for improving language and readability. After using this tool/service,
the authors reviewed and edited the content as needed and take full
responsibility for the content of the published article.

\end{document}